\documentclass{article}

\usepackage{microtype}
\usepackage{graphicx}
\usepackage{subcaption}
\usepackage{booktabs} 
\usepackage{hyperref}

\usepackage{tabularx,ragged2e}
\usepackage{booktabs}
\newcolumntype{L}{>{\centering\arraybackslash}X}

\usepackage{multicol}
\usepackage{multirow}

\usepackage{boldline}

\usepackage[]{graphicx}

\usepackage{amssymb}
\usepackage{pifont}
\newcommand{\cmark}{\ding{51}}%
\newcommand{\xmark}{\ding{55}}%

\usepackage[accepted]{icml2024}

\usepackage{amsmath}
\usepackage{amssymb}
\usepackage{mathtools}
\usepackage{amsthm}

\usepackage[capitalize,noabbrev]{cleveref}

\theoremstyle{plain}

\theoremstyle{definition}

\theoremstyle{remark}

\usepackage[textsize=tiny]{todonotes}

\icmltitlerunning{Training-Free ViT Acceleration with Delayed Spatial Merging}

\begin{document}

\twocolumn[

\icmltitle{Training-Free Acceleration of ViTs with Delayed Spatial Merging}




\begin{icmlauthorlist}
\icmlauthor{Jung Hwan Heo}{yyy}
\icmlauthor{Seyedarmin Azizi}{yyy}
\icmlauthor{Arash Fayyazi}{yyy}
\icmlauthor{Massoud Pedram}{yyy}
\end{icmlauthorlist}

\icmlaffiliation{yyy}{University of Southern California (USC)}

\icmlcorrespondingauthor{Jung Hwan Heo}{jhjohnheo@gmail.com}

\icmlkeywords{Machine Learning, ICML}

\vskip 0.3in
]

\printAffiliationsAndNotice{}  

\begin{abstract}
Token merging has emerged as a new paradigm that can accelerate the inference of Vision Transformers (ViTs)
without any retraining or finetuning.
To push the frontier of \textbf{training-free} acceleration in ViTs, we improve token merging by adding the perspectives of 1) activation outliers and 2) hierarchical representations.
Through a careful analysis of the attention behavior in ViTs, we characterize a delayed onset of the \textit{convergent attention phenomenon}, which makes token merging undesirable in the bottom blocks of ViTs.
Moreover, we augment token merging with a hierarchical processing scheme to capture \textit{multi-scale redundancy} between visual tokens.
Combining these two insights, we
build a unified inference framework called \textbf{DSM}: \textbf{D}elayed \textbf{S}patial \textbf{M}erging.
We extensively evaluate DSM on various ViT model scales (Tiny to Huge) and tasks (ImageNet-1k and transfer learning), achieving up to 1.8$\times$ FLOP reduction and 1.6$\times$ throughput speedup at a negligible loss while being \textit{two orders of magnitude faster} than existing methods.
\end{abstract}

\section{Introduction}
Transformers \citep{vaswani2017attention} has become a general-purpose backbone architecture that drove great progress in language modeling \citep{DBLP:conf/naacl/DevlinCLT19}, speech recognition \citep{DBLP:conf/icassp/TianYBTZW20}, to computer vision \citep{dosovitskiy2020image}.
Compared to Convolutional Neural Networks (CNNs), Vision Transformers (ViTs) have minimal inductive bias, benefiting from large-scale pretraining.
Modern self-supervised models such as MAE obtain up to 90.94\% top-1 accuracy on ImageNet-1k \citep{pmlr-v162-wortsman22a}. 

However, efficient deployment of ViTs remains a challenge due to the large model size.
A major line of work has focused on pruning task-irrelevant tokens with various importance metrics such as token embeddings \citep{yin2022vit}, attention scores \citep{liang2022expediting}, and lightweight neural network predictors \citep{rao2021dynamicvit}.
Recently, a newly proposed token merging scheme enabled a \emph{training-free} approach to token reduction \citep{bolya2022tome}.
While prior work can effectively accelerate ViT inference, using these techniques in practice is still challenging. Previous approaches train from scratch \citep{liang2022expediting}, fine-tune with extra parameters \citep{rao2021dynamicvit}, and optimize with additional loss functions that increase the wall-clock training time \citep{yin2022vit}. Such complexities introduce extra computational budgets and engineering efforts that prevent the easy adoption of techniques. Token merging scheme can avoid this via the training-free mode \citep{bolya2022tome}, but it incurs nontrivial accuracy loss, which ultimately necessitates training from scratch to achieve competitive performance. 

To push the frontier of training-free ViT acceleration via token merging, we turn to recent findings of activation outliers in large Transformers~\citep{darcet2023registers,xiao2023streamingllm} as well as a principled hierarchical processing technique~\citep{jarrett2009multi, lee2009heirarchical, krizhevsky2009learning}.
Augmented by our delayed and hierarchical merging schemes, DSM yields a strong token merging technique that is aware of both Transformer attention mechanics and multi-scale redundancies.
Our contributions are summarized:
\begin{itemize}
    \item We find that the recently discovered high-norm token outliers in ViTs~\citep{darcet2023registers} are attributed to the Attention Sink behavior in language models~\citep{xiao2023streamingllm}. By carefully studying the attention mechanics in ViTs, we identify an intriguing phenomenon that we call \textit{delayed convergent attention}.
    \item Motivated by the observation that 1) token merging is undesirable in the bottom Transformer blocks and 2) hierarchical image processing captures multi-scale interactions, we present a unified inference framework called Delayed Spatial Merging (DSM).
    \item We extensively evaluate DSM on ViT and DeiT models of various scales (Tiny $\sim$ Huge) on ImageNet-1k and transfer learning tasks. With no more than a 1\% drop in accuracy, our framework achieves 1.8$\times$ FLOP reduction and 1.6$\times$ speedup on NVIDIA A6000 GPU.
\end{itemize}

\begin{figure*}[ht]
\centering
\includegraphics[width=0.8\paperwidth]{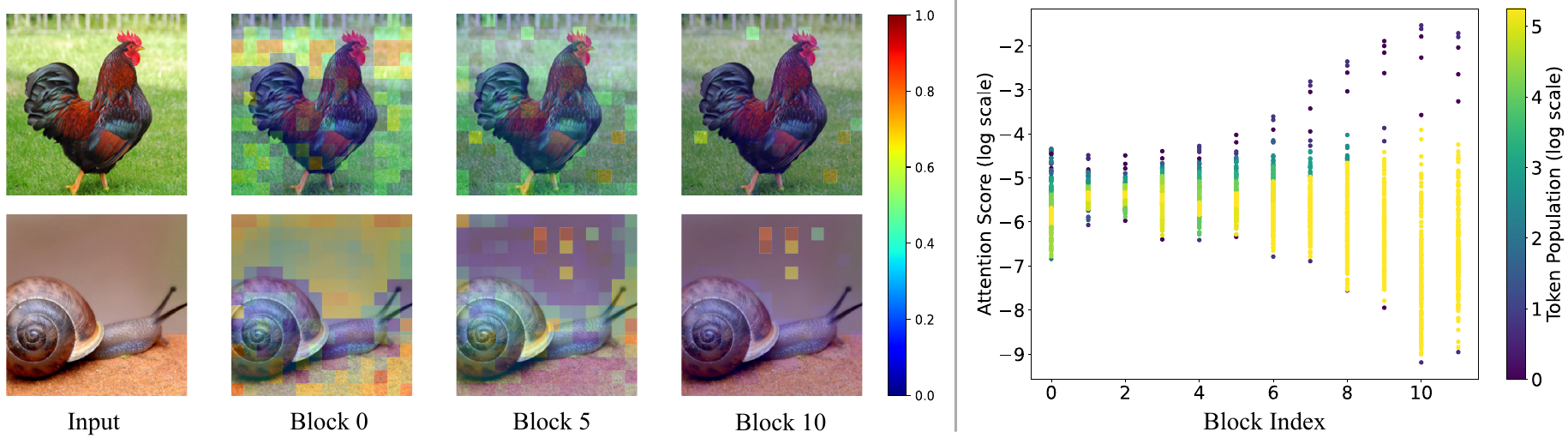}
\vspace{-1em}
\caption{
\textbf{Connection between high-norm activation outliers and Attention Sinks on ViT-S.}
\textit{Left:} Low-information background tokens progressively collect most of the attention scores.
\textit{Right:} Such outlier tokens receive orders of magnitude higher attention values. Critically, we observe the \textit{delay} in which the outlier tokens begin to emerge, which inspires further investigation of the attention behavior.
}
\label{fig:attention-sink}
\end{figure*}

\section{Delayed Spatial Merging} \label{sec:framework}

\paragraph{Tracing Attention Sinks in ViTs.} 

Recently, high-norm activation outliers have been observed in ViTs, which act as registers that pool global information~\citep{darcet2023registers,bondarenko2023quantizable}. We find inspiration from the Attention Sink behavior from language models~\citep{xiao2023streamingllm} to trace the source of such outliers.
As in Figure~\ref{fig:attention-sink}, we first verify that the high-norm outlier tokens in ViTs are related to the Attention Sink behavior.
Although initialized according to a nearly uniform distribution, attention scores are progressively accumulated on only a few background tokens, leading to orders of magnitude differences between scores of the outlier sink tokens and the other token.
Interestingly, we observe that there is an initial delay before the attention sinks begin to emerge.
This naturally raises two questions: \textit{Why does this delay exist, and how does it affect token merging?}

\begin{figure}[t!]
\centering
\includegraphics[width=\linewidth]{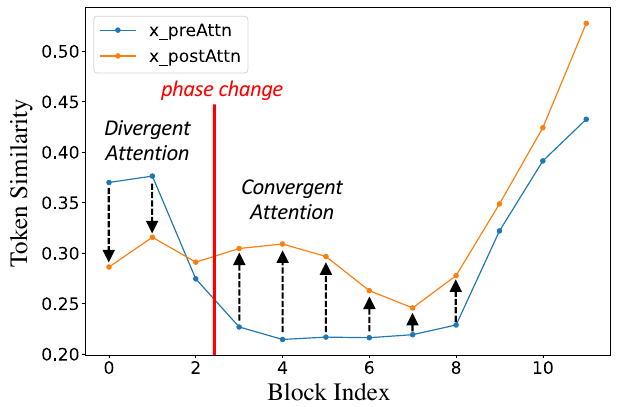}
\vspace{-2em}
\caption{
\textbf{Illustration of Delayed Convergent Attention.} 
The first few attention blocks have decreasing similarity (computed with Equation~\ref{eq:toksim} on DeiT-S) and then it increases for the rest of the network. It is desirable to merge tokens when they are becoming similar (convergent), motivating the delayed merging scheme.
}
\vspace{-1em}
\label{fig:show}
\end{figure}

\subsection{Delayed Merging} 
\paragraph{Vanilla Token Merging.}
Transformer block in a ViT consists of a multi-head attention ($\mathrm{MHA}$) layer and a Feedforward Network ($\mathrm{FFN}$) layer. For the $l$-th transformer block in a network of depth $L$, the forward pass is expressed as

\vspace{-1em}
\begin{equation} \label{eq:fw_pass}
    \bar{\mathbf{X}}^l = \mathbf{X}^l + \mathrm{MHA}(\mathbf{X}^{l}),\\
    \mathbf{X}^{l+1} = \bar{\mathbf{X}}^{l} + \mathrm{FFN}(\bar{\mathbf{X}}^{l}),
\end{equation}
where $\mathbf{X}^l\in \mathbb{R}^{N\times C}$ is the input sequence with $N$ tokens, each with an embedding size of $C$.
Token merging is applied within each transformer block between $\mathrm{MHA}$ and $\mathrm{FFN}$.
Given a sequence of $n$ tokens ($\mathrm{MHA}$ layer output), denoted by $\bar{\mathbf{X}}^l = [x_1, ..., x_n] $, a weighted complete bipartite graph comprising two sets of nodes (tokens): $\mathbb{A} = [x_1, x_3, ..., x_{n-1}]$ and $\mathbb{B} = [x_2, x_4, ..., x_n]$ is constructed.
An edge between token $a \in \mathbb{A}$ and token $b \in \mathbb{B}$ captures the cosine similarity between embeddings of $a$ and $b$.
A weighted bipartite graph matching algorithm is then applied to identify the set of $r \leq n/2$ edges that have the maximum weighted sum.
The tokens associated with each of these $r$ edges are merged using a channel-wise weighted average.
Finally, the two sets $\mathbb{A}$ and $\mathbb{B}$ are combined to yield a truncated sequence of tokens with $r$ fewer tokens. 

\begin{figure*}[t!]
\centering
\includegraphics[width=0.7\linewidth]{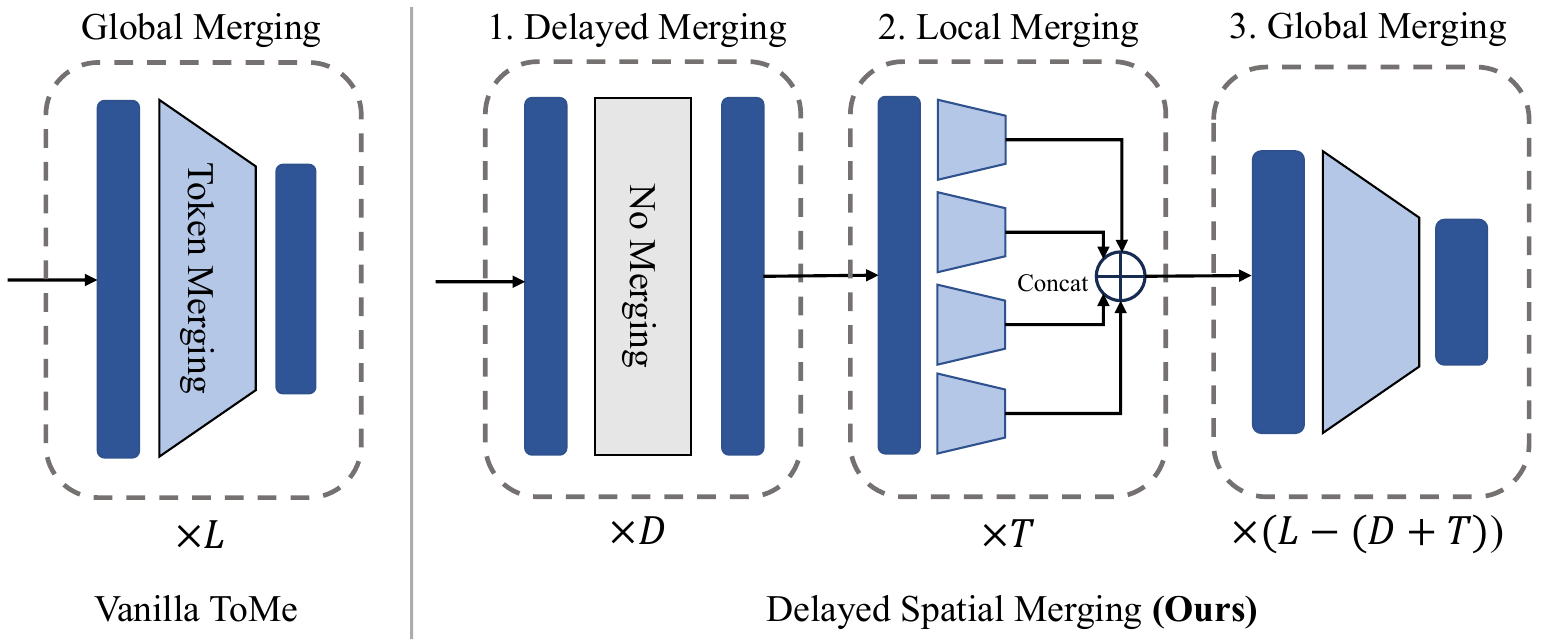}
\caption{
\textbf{Left:} The original Token Merging (ToMe) by \citeauthor{bolya2022tome} is globally applied to all tokens for all $L$ transformer blocks. \textbf{Right:} Motivated by the principles of convergent attention and spatial awareness, our proposed Delayed Spatial Merging (DSM) augments ToMe by not merging in the initial $D$ blocks, locally merging for $T$ blocks, then globally merging for the rest of the network. }
\label{fig:dsm}
\end{figure*}

\paragraph{Characterizing Convergent Attention.}
We now investigate how the delay in Attention Sinks affects token similarity distribution.
Intuitively, an ideal scenario to conduct token merging would be when tokens are most similar to each other (i.e., avoid forced merging of tokens when they are dissimilar.) 
To quantify the degree of similarity among tokens, we adopt the token similarity metric which has been widely used in text generation~\citep{zhang2019bertscore}:
\begin{equation} \label{eq:toksim}
Sim = \frac{1}{n\left(n-1\right)} \sum_{i \neq j} \frac{x_i^T  x_j}{\vert \vert x_i\vert \vert_2  \vert \vert x_j \vert \vert_2} ,  
\end{equation}

A higher $Sim$ score indicates that the tokens in the layer have similar embeddings.
Interestingly as in Figure~\ref{fig:show}, the initial blocks have tokens become less similar during attention (\textit{divergent} attention) while after a certain point in the ViT (a phase change starting at block 2), tokens consistently become more similar (\textit{convergent} attention). 
As merging tokens that are in the process of diversifying is counterproductive, we delay merging until token embeddings stabilize to exhibit the convergent attention behavior. 

\subsection{Spatial Merging} \label{subsec:lgtm}
Hierarchical image processing is a fundamental technique that spans a wide range of computer vision modeling from semantic segmentation~\citep{long2015fully}, object detection~\citep{jarrett2009multi,lin2017feature}, to 3D rendering via Neural Radiance Fields~\citep{barron2021mip}. We introduce the principle of hierarchical representations to token merging for the first time. The intuition is to capture multi-scale interactions between visual tokens such that the similarity (feature redundancy) search process can be done in finer granularity.

Neighboring pixels in an image having stronger semantic relationships with each other; for example, a picture of an animal has contiguous body parts where \emph{spatial proximity} correlates well with \emph{semantic similarity}.
Instead of globally searching for similar tokens, we constrain the search space to local windows.
The input tokens can be represented as a 2D grid with dimensions (H, W), which we partition into four equally-sized windows with dimension $w$.
To minimize the complexity, we set initial window size to $w=7$ in all of our experiments as it nicely divides 14$\times$14 grid of tokens (224$\times$224 resolution w/ common patch size of $16$).
When the number of tokens is not divisible by $w$, we apply padding in the bottom right to retain the 2D formation.

Rather than a static window size $w$, we progressively increment the window size in every block.
This is based on the intuition that positional similarity, as positional embeddings are added right before block 0, is most relevant in the earlier part of the network.
Thus, we increment the windows every block until it equivalently reduces to global merging (where the window is as big as the remaining 2D grid of tokens).
Windows can be stacked to efficiently merge tokens in parallel. This is possible because token merging is applied independently for each example in a batch, and the window dimension can be fused into the batch dimension B:
(B, H, W) $\rightarrow$ (B, H/$w$, $w$, W/$w$, $w$)
$\rightarrow$ (B * H/$w$ * W/$w$, $w$, $w$). Efficient kernel implementation of the window stack operation is possible as demonstrated in~\cite{liu2021swin}.

\subsection{Unified Inference Framework} \label{subsec:overall}
As in \cref{fig:dsm}, DSM augments the vanilla token merging technique with delayed merging and localized merging. For a network with depth $L$, we delay for $D$ blocks, apply localized merging for $T$ blocks with a window size of $w$, and execute global merging for the rest of the network. The only hyperparameter we tune is $r$, which is the number of tokens to reduce in a single token merging layer; we further discuss hyperparameter settings in Section~\ref{appdx:method}.
\vspace{-1em}
\begin{table}[!b]
    \centering
    \caption{\textbf{Comparison to Prior Work.}  Our framework provides competitive performance while being two orders of magnitude faster. E2E training time is measured in a single 8 GPU node.
    }
    \label{tab:results-training}
    \resizebox{\columnwidth}{!}{
    \begin{tabular}{lcccc}
    \toprule
    {}       & Top-1 & GFLOP & Epochs & E2E (hrs) \\
    \midrule
    DeiT-S    & 79.8  & 4.6 & 0     & 0   \\
    \midrule
    DynamicViT~\cite{rao2021dynamicvit}  & 79.3  & 2.9 & 30     & 44.8    \\
    SPViT~\cite{kong2021spvit}       & 79.3  & 2.6 & 60     & --      \\
    A-ViT~\cite{yin2022vit}       & 78.6  & 2.9 & 100    & 76.4    \\
    E-ViT~\cite{liang2022expediting}       & 79.1  & 2.6 & 300    & 154.4    \\
    ATS~\cite{fayyaz2022ats}         & 79.7  & 2.9 & 30     & --      \\
    ToMe~\cite{bolya2022tome}  & 79.4  & 2.7  & 300    & 102.2    \\
    \midrule
    Spatial Merging \textbf{(Ours)} & 79.3 & 2.8 & 0   & 0   \\
    DSM \textbf{(Ours)} & 78.6 & 2.5 & 0   & 0    \\
    \bottomrule
    \end{tabular}
}
\end{table}

\begin{figure*}[t!]
\centering
\includegraphics[width=0.8\linewidth]{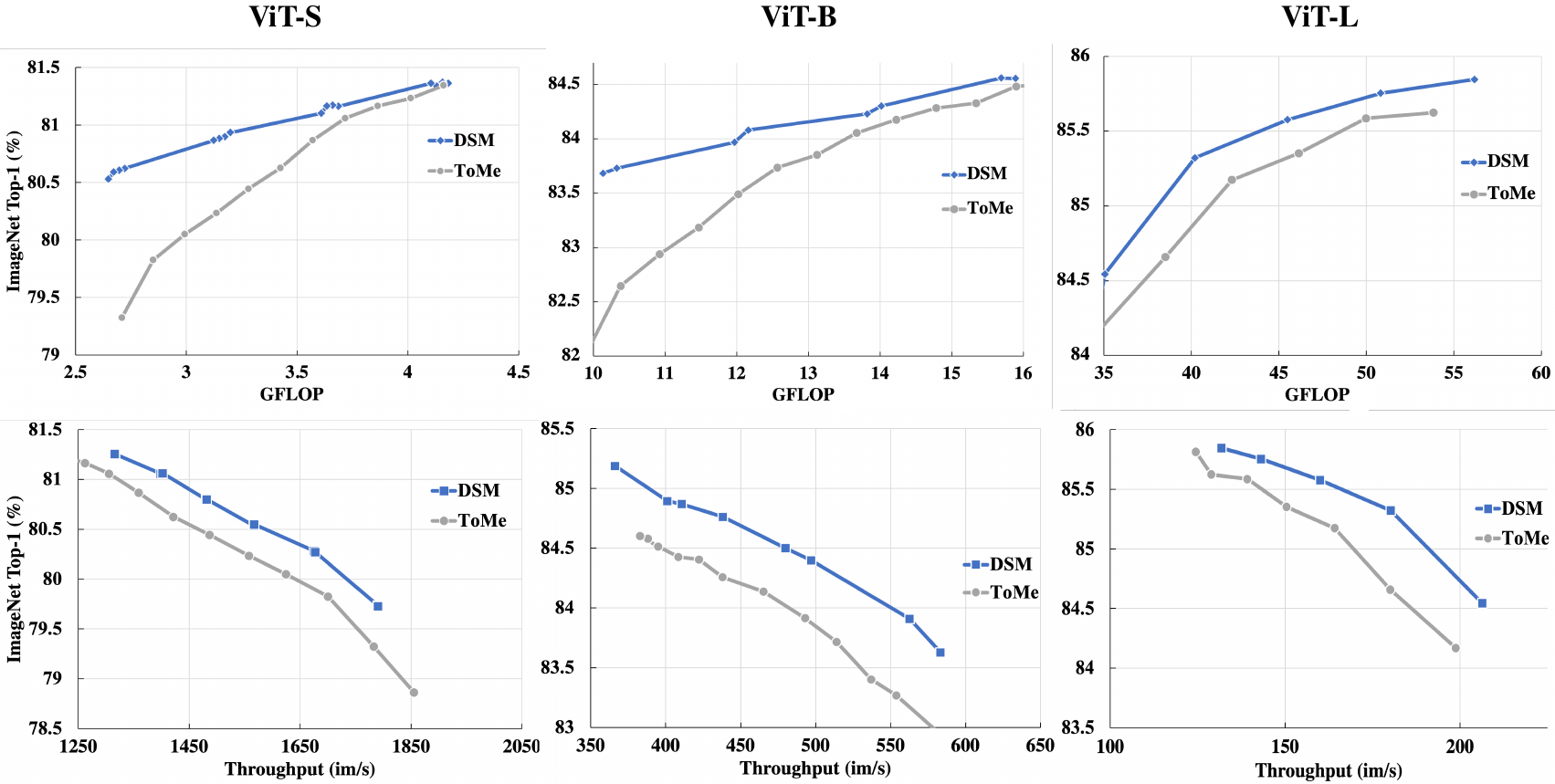}
\caption{\textbf{Model Sweep.} We apply our inference framework to several state-of-the-art ViT models in a \emph{training-free} fashion. DSM's hyperparameters are fixed via network architecture, only varying parameter \emph{r} to produce Top-1 Acc. vs. GFLOP curves on ImageNet-1k.}
\label{fig:vit-sweeps}
\end{figure*}

\begin{table*}[t!]
    \small
    \centering
    \resizebox{0.8\linewidth}{!}{
    \begin{tabular}{lcccccccccccccc}
        \hline
        \multirow{14}{*}{} \\
         \multirow{14}{*}{} & \multicolumn{3}{c}{Oxford-IIIT Pet} & \multicolumn{3}{c}{Flowers102} & \multicolumn{3}{c}{FGVC-Aircraft} & \multicolumn{3}{c}{CIFAR-100} \\
        \cmidrule(lr){2-4}  
        \cmidrule(lr){5-7}
        \cmidrule(lr){8-10}
        \cmidrule(lr){11-13}
        & acc@1 & GFLOP & im/s & acc@1 & GFLOP & im/s & acc@1 & GFLOP & im/s & acc@1 & GFLOP & im/s \\
        \midrule \midrule
        Baseline & 92.12 & 17.57 & 404.53 &98.13 &17.57 &407.27 &  81.12 & 17.57& 408.88 & 91.10 & 17.57 & 402.51\\
        \midrule
        $r=4$ & 92.04 & 16.10 & 384.51 & 98.19 & 16.10 & 384.54 & 80.98& 16.10 & 388.25 & 90.96 & 16.10 & 383.1  \\
        $r=8$ & 92.01 & 14.46 & 431.37 & 97.95 & 14.46 & 430.18 & 80.95  & 14.46 & 436.20 & 91.01 & 14.46 & 431.72 \\
        $r=12$ & 91.74 & 13.27 & 468.78 & 97.69 & 13.27 & 466.17 & 81.07 & 13.27 & 471.10 & 90.90 & 13.27 & 469.10\\
        $r=16$ & 91.55 & 11.96& 517.05& 97.45 & 11.96 & 513.75 & 80.38 & 11.96 & 518.86 & 90.76 & 11.96 & 517.76\\
        $r=20$ & 91.32 &10.16 & 612.50& 97.14 & 10.16 & 609.61 & 80.20 & 10.16 & 612.11 & 90.25 & 10.16 & 613.92 \\
        \hline
    \end{tabular}
    }
    \vspace{1em}
    \caption{\textbf{Transfer Learning.} Fine-tuned ViT-B accelerated with DSM consistently achieves 1.5$\times$ speedup across various datasets.}
    \label{results:transfer}
\end{table*}

\section{Experiments}

We conduct our experiments on ImageNet-1K~\cite{russakovsky2015imagenet} to evaluate the effectiveness of our method in accelerating off-the-shelf ViTs on classification tasks. Both DeiT~\cite{DBLP:conf/icml/TouvronCDMSJ21} and ViT models trained with AugReg~\cite{DBLP:journals/corr/abs-2106-10270} are used to test the generalizability of our method across different backbones and training methods.
The computational cost is measured in FLOP with the Torchprofiler\footnote{https://github.com/zhijian-liu/torchprofile} library. Inference throughput is measured on an Nvidia RTX A6000 GPU with a fixed batch size of 32 averaged over 50 runs.

As in \cref{tab:results-training}, our DSM achieves competitive performance while being \emph{two orders of magnitude faster} than existing approaches thanks to the training-free approach. For example, E-ViT \cite{liang2022expediting} takes around 154 single GPU hours for one run. Since it requires running the method for each target speedup, the cost of deploying to various resource constraints can become quickly intractable.

In \cref{fig:vit-sweeps}, we apply our framework to ViT-[S, B, L] \emph{off-the-shelf}with 224px and patch size 16. For each model, we benchmark DSM against ToMe. We vary $r$ to construct two Pareto curves that compare Top-1 accuracy to \#MACs and throughput. Note, we sweep with higher $r$ values with the DSM to match the computational load of ToMe.

We can see that our framework consistently gives better results than ToMe, especially for smaller models.
Remarkably, we can save 45\% and 42\% of the FLOP within a 1\% loss for ViT-S and ViT-B, respectively.
Relative to vanilla ToMe, it can improve the accuracy by more than 1\%. Yet, the success of DSM inversely scales with model size, showing a negligible gain for ViT-L.
Compared to the success of DSM in saving FLOP, the throughput gains are relatively marginal.
We think this is because the additional data movements, such as sorting, padding, and modifying tensor dimensions, cause a nontrivial overhead. 
Interestingly, this overhead is less obvious for larger models, as the DSM curve shifts to the right with better trade-off margins.
With larger models that executes heavy loads of matrix multiplications, computing becomes a bottleneck rather than data movement.
This makes the memory I/O overhead from localized marging less evident for larger models.




\nocite{langley00}

\bibliography{main}

\newcommand{\noop}[1]{}
\begin{thebibliography}{44}
\providecommand{\natexlab}[1]{#1}
\providecommand{\url}[1]{\texttt{#1}}
\expandafter\ifx\csname urlstyle\endcsname\relax
  \providecommand{\doi}[1]{doi: #1}\else
  \providecommand{\doi}{doi: \begingroup \urlstyle{rm}\Url}\fi

\bibitem[Barron et~al.(2021)Barron, Mildenhall, Tancik, Hedman, Martin-Brualla, and Srinivasan]{barron2021mip}
Barron, J.~T., Mildenhall, B., Tancik, M., Hedman, P., Martin-Brualla, R., and Srinivasan, P.~P.
\newblock Mip-nerf: A multiscale representation for anti-aliasing neural radiance fields.
\newblock In \emph{Proceedings of the IEEE/CVF International Conference on Computer Vision}, pp.\  5855--5864, 2021.

\bibitem[Beyer et~al.(2022)Beyer, Zhai, Royer, Markeeva, Anil, and Kolesnikov]{beyer2022knowledge}
Beyer, L., Zhai, X., Royer, A., Markeeva, L., Anil, R., and Kolesnikov, A.
\newblock Knowledge distillation: A good teacher is patient and consistent.
\newblock In \emph{Proceedings of the IEEE/CVF Conference on Computer Vision and Pattern Recognition}, pp.\  10925--10934, 2022.

\bibitem[Bolya et~al.(2023)Bolya, Fu, Dai, Zhang, Feichtenhofer, and Hoffman]{bolya2022tome}
Bolya, D., Fu, C.-Y., Dai, X., Zhang, P., Feichtenhofer, C., and Hoffman, J.
\newblock Token merging: Your {ViT} but faster.
\newblock In \emph{International Conference on Learning Representations}, 2023.

\bibitem[Bondarenko et~al.(2023)Bondarenko, Nagel, and Blankevoort]{bondarenko2023quantizable}
Bondarenko, Y., Nagel, M., and Blankevoort, T.
\newblock Quantizable transformers: Removing outliers by helping attention heads do nothing.
\newblock \emph{arXiv preprint arXiv:2306.12929}, 2023.

\bibitem[Darcet et~al.(2023)Darcet, Oquab, Mairal, and Bojanowski]{darcet2023registers}
Darcet, T., Oquab, M., Mairal, J., and Bojanowski, P.
\newblock Vision transformers need registers.
\newblock \emph{arXiv preprint arXiv:2309.16588}, 2023.

\bibitem[Dettmers et~al.(2022)Dettmers, Lewis, Belkada, and Zettlemoyer]{dettmers2022llm}
Dettmers, T., Lewis, M., Belkada, Y., and Zettlemoyer, L.
\newblock Llm. int8 (): 8-bit matrix multiplication for transformers at scale.
\newblock \emph{arXiv preprint arXiv:2208.07339}, 2022.

\bibitem[Devlin et~al.(2019)]{DBLP:conf/naacl/DevlinCLT19}
Devlin, J. et~al.
\newblock {BERT:} pre-training of deep bidirectional transformers for language understanding.
\newblock In \emph{Conference of the North American Chapter of the Association for Computational Linguistics: Human Language Technologies}. Association for Computational Linguistics, 2019.

\bibitem[Dosovitskiy et~al.(2020)Dosovitskiy, Beyer, Kolesnikov, Weissenborn, Zhai, Unterthiner, Dehghani, Minderer, Heigold, Gelly, et~al.]{dosovitskiy2020image}
Dosovitskiy, A., Beyer, L., Kolesnikov, A., Weissenborn, D., Zhai, X., Unterthiner, T., Dehghani, M., Minderer, M., Heigold, G., Gelly, S., et~al.
\newblock An image is worth 16x16 words: Transformers for image recognition at scale.
\newblock \emph{arXiv preprint arXiv:2010.11929}, 2020.

\bibitem[Fayyaz et~al.(2022)Fayyaz, Koohpayegani, Jafari, Sengupta, Joze, Sommerlade, Pirsiavash, and Gall]{fayyaz2022ats}
Fayyaz, M., Koohpayegani, S.~A., Jafari, F.~R., Sengupta, S., Joze, H. R.~V., Sommerlade, E., Pirsiavash, H., and Gall, J.
\newblock Adaptive token sampling for efficient vision transformers.
\newblock In \emph{European Conference on Computer Vision}, pp.\  396--414. Springer, 2022.

\bibitem[Han et~al.(2015)Han, Mao, and Dally]{han2015deep}
Han, S., Mao, H., and Dally, W.~J.
\newblock Deep compression: Compressing deep neural networks with pruning, trained quantization and huffman coding.
\newblock \emph{arXiv preprint arXiv:1510.00149}, 2015.

\bibitem[He et~al.(2016)He, Zhang, Ren, and Sun]{he2016deep}
He, K., Zhang, X., Ren, S., and Sun, J.
\newblock Deep residual learning for image recognition.
\newblock In \emph{Proceedings of the IEEE conference on computer vision and pattern recognition}, pp.\  770--778, 2016.

\bibitem[Heo et~al.(2023)Heo, Kim, Kwon, Kim, Kwon, and Lee]{heo2023rethinking}
Heo, J.~H., Kim, J., Kwon, B., Kim, B., Kwon, S.~J., and Lee, D.
\newblock Rethinking channel dimensions to isolate outliers for low-bit weight quantization of large language models.
\newblock \emph{arXiv preprint arXiv:2309.15531}, 2023.

\bibitem[Hinton et~al.(2015)Hinton, Vinyals, and Dean]{DBLP:journals/corr/HintonVD15}
Hinton, G.~E., Vinyals, O., and Dean, J.
\newblock Distilling the knowledge in a neural network.
\newblock \emph{CoRR}, abs/1503.02531, 2015.

\bibitem[Jarrett et~al.(2009)Jarrett, Kavukcuoglu, Ranzato, and LeCun]{jarrett2009multi}
Jarrett, K., Kavukcuoglu, K., Ranzato, M., and LeCun, Y.
\newblock What is the best multi-stage architecture for object recognition?
\newblock In \emph{2009 IEEE 12th international conference on computer vision}, pp.\  2146--2153. IEEE, 2009.

\bibitem[Kim et~al.(2021)Kim, Gholami, Yao, Mahoney, and Keutzer]{DBLP:conf/icml/KimGYMK21}
Kim, S., Gholami, A., Yao, Z., Mahoney, M.~W., and Keutzer, K.
\newblock {I-BERT:} integer-only {BERT} quantization.
\newblock In Meila, M. and Zhang, T. (eds.), \emph{Proceedings of the 38th International Conference on Machine Learning, {ICML} 2021, 18-24 July 2021, Virtual Event}, volume 139 of \emph{Proceedings of Machine Learning Research}, pp.\  5506--5518. {PMLR}, 2021.
\newblock URL \url{http://proceedings.mlr.press/v139/kim21d.html}.

\bibitem[Kong et~al.(2021)Kong, Dong, Ma, Meng, Niu, Sun, Ren, Qin, Tang, and Wang]{kong2021spvit}
Kong, Z., Dong, P., Ma, X., Meng, X., Niu, W., Sun, M., Ren, B., Qin, M., Tang, H., and Wang, Y.
\newblock Spvit: Enabling faster vision transformers via soft token pruning.
\newblock \emph{arXiv preprint arXiv:2112.13890}, 2021.

\bibitem[Kong et~al.(2022)Kong, Dong, Ma, Meng, Niu, Sun, Shen, Yuan, Ren, Tang, Qin, and Wang]{10.1007/978-3-031-20083-0_37}
Kong, Z., Dong, P., Ma, X., Meng, X., Niu, W., Sun, M., Shen, X., Yuan, G., Ren, B., Tang, H., Qin, M., and Wang, Y.
\newblock Spvit: Enabling faster vision transformers via latency-aware soft token pruning.
\newblock In Avidan, S., Brostow, G., Ciss{\'e}, M., Farinella, G.~M., and Hassner, T. (eds.), \emph{Computer Vision -- ECCV 2022}, pp.\  620--640, Cham, 2022. Springer Nature Switzerland.

\bibitem[Krizhevsky et~al.(2009)Krizhevsky, Hinton, et~al.]{krizhevsky2009learning}
Krizhevsky, A., Hinton, G., et~al.
\newblock Learning multiple layers of features from tiny images.
\newblock 2009.

\bibitem[Kurtic et~al.(2022)Kurtic, Campos, Nguyen, Frantar, Kurtz, Fineran, Goin, and Alistarh]{DBLP:journals/corr/abs-2203-07259}
Kurtic, E., Campos, D., Nguyen, T., Frantar, E., Kurtz, M., Fineran, B., Goin, M., and Alistarh, D.
\newblock The optimal {BERT} surgeon: Scalable and accurate second-order pruning for large language models.
\newblock \emph{CoRR}, abs/2203.07259, 2022.
\newblock \doi{10.48550/arXiv.2203.07259}.
\newblock URL \url{https://doi.org/10.48550/arXiv.2203.07259}.

\bibitem[Langley(2000)]{langley00}
Langley, P.
\newblock Crafting papers on machine learning.
\newblock In Langley, P. (ed.), \emph{Proceedings of the 17th International Conference on Machine Learning (ICML 2000)}, pp.\  1207--1216, Stanford, CA, 2000. Morgan Kaufmann.

\bibitem[Lee et~al.(2009)Lee, Grosse, Ranganath, and Ng]{lee2009heirarchical}
Lee, H., Grosse, R., Ranganath, R., and Ng, A.~Y.
\newblock Convolutional deep belief networks for scalable unsupervised learning of hierarchical representations.
\newblock In \emph{Proceedings of the 26th annual international conference on machine learning}, pp.\  609--616, 2009.

\bibitem[Liang et~al.(2022)Liang, Yuan, Ding, Luo, Lin, Jia, Zhang, Zhang, and Hu]{liang2022expediting}
Liang, W., Yuan, Y., Ding, H., Luo, X., Lin, W., Jia, D., Zhang, Z., Zhang, C., and Hu, H.
\newblock Expediting large-scale vision transformer for dense prediction without fine-tuning.
\newblock \emph{Advances in Neural Information Processing Systems}, 35:\penalty0 35462--35477, 2022.

\bibitem[Lin et~al.(2023)Lin, Tang, Tang, Yang, Dang, and Han]{lin2023awq}
Lin, J., Tang, J., Tang, H., Yang, S., Dang, X., and Han, S.
\newblock Awq: Activation-aware weight quantization for llm compression and acceleration.
\newblock \emph{arXiv preprint arXiv:2306.00978}, 2023.

\bibitem[Lin et~al.(2017)Lin, Doll{\'a}r, Girshick, He, Hariharan, and Belongie]{lin2017feature}
Lin, T.-Y., Doll{\'a}r, P., Girshick, R., He, K., Hariharan, B., and Belongie, S.
\newblock Feature pyramid networks for object detection.
\newblock In \emph{Proceedings of the IEEE conference on computer vision and pattern recognition}, pp.\  2117--2125, 2017.

\bibitem[Liu et~al.(2021)Liu, Lin, Cao, Hu, Wei, Zhang, Lin, and Guo]{liu2021swin}
Liu, Z., Lin, Y., Cao, Y., Hu, H., Wei, Y., Zhang, Z., Lin, S., and Guo, B.
\newblock Swin transformer: Hierarchical vision transformer using shifted windows.
\newblock In \emph{Proceedings of the IEEE/CVF International Conference on Computer Vision}, pp.\  10012--10022, 2021.

\bibitem[Long et~al.(2015)Long, Shelhamer, and Darrell]{long2015fully}
Long, J., Shelhamer, E., and Darrell, T.
\newblock Fully convolutional networks for semantic segmentation.
\newblock In \emph{Proceedings of the IEEE conference on computer vision and pattern recognition}, pp.\  3431--3440, 2015.

\bibitem[Marin et~al.(2021)Marin, Chang, Ranjan, Prabhu, Rastegari, and Tuzel]{DBLP:journals/corr/abs-2110-03860}
Marin, D., Chang, J.~R., Ranjan, A., Prabhu, A., Rastegari, M., and Tuzel, O.
\newblock Token pooling in vision transformers.
\newblock \emph{CoRR}, abs/2110.03860, 2021.
\newblock URL \url{https://arxiv.org/abs/2110.03860}.

\bibitem[Miller(2023)]{softmax1}
Miller, E.
\newblock Attention is off by one.
\newblock 2023.
\newblock URL \url{https://www.evanmiller.org/attention-is-off-by-one.html}.

\bibitem[Radosavovic et~al.(2020)Radosavovic, Kosaraju, Girshick, He, and Doll{\'a}r]{radosavovic2020regnet}
Radosavovic, I., Kosaraju, R.~P., Girshick, R., He, K., and Doll{\'a}r, P.
\newblock Designing network design spaces.
\newblock In \emph{Proceedings of the IEEE/CVF conference on computer vision and pattern recognition}, pp.\  10428--10436, 2020.

\bibitem[Rao et~al.(2021)Rao, Zhao, Liu, Lu, Zhou, and Hsieh]{rao2021dynamicvit}
Rao, Y., Zhao, W., Liu, B., Lu, J., Zhou, J., and Hsieh, C.-J.
\newblock Dynamicvit: Efficient vision transformers with dynamic token sparsification.
\newblock \emph{Advances in neural information processing systems}, 34:\penalty0 13937--13949, 2021.

\bibitem[Russakovsky et~al.(2015)Russakovsky, Deng, Su, Krause, Satheesh, Ma, Huang, Karpathy, Khosla, Bernstein, et~al.]{russakovsky2015imagenet}
Russakovsky, O., Deng, J., Su, H., Krause, J., Satheesh, S., Ma, S., Huang, Z., Karpathy, A., Khosla, A., Bernstein, M., et~al.
\newblock Imagenet large scale visual recognition challenge.
\newblock \emph{International journal of computer vision}, 115:\penalty0 211--252, 2015.

\bibitem[Ryoo et~al.(2021)Ryoo, Piergiovanni, Arnab, Dehghani, and Angelova]{DBLP:conf/nips/RyooPADA21}
Ryoo, M.~S., Piergiovanni, A.~J., Arnab, A., Dehghani, M., and Angelova, A.
\newblock Tokenlearner: Adaptive space-time tokenization for videos.
\newblock In Ranzato, M., Beygelzimer, A., Dauphin, Y.~N., Liang, P., and Vaughan, J.~W. (eds.), \emph{Advances in Neural Information Processing Systems 34: Annual Conference on Neural Information Processing Systems 2021, NeurIPS 2021, December 6-14, 2021, virtual}, pp.\  12786--12797, 2021.
\newblock URL \url{https://proceedings.neurips.cc/paper/2021/hash/6a30e32e56fce5cf381895dfe6ca7b6f-Abstract.html}.

\bibitem[Shen et~al.(2020)Shen, Dong, Ye, Ma, Yao, Gholami, Mahoney, and Keutzer]{DBLP:conf/aaai/ShenDYMYGMK20}
Shen, S., Dong, Z., Ye, J., Ma, L., Yao, Z., Gholami, A., Mahoney, M.~W., and Keutzer, K.
\newblock {Q-BERT:} hessian based ultra low precision quantization of {BERT}.
\newblock In \emph{The Thirty-Fourth {AAAI} Conference on Artificial Intelligence, {AAAI} 2020, The Thirty-Second Innovative Applications of Artificial Intelligence Conference, {IAAI} 2020, The Tenth {AAAI} Symposium on Educational Advances in Artificial Intelligence, {EAAI} 2020, New York, NY, USA, February 7-12, 2020}, pp.\  8815--8821. {AAAI} Press, 2020.
\newblock URL \url{https://ojs.aaai.org/index.php/AAAI/article/view/6409}.

\bibitem[Steiner et~al.(2021)Steiner, Kolesnikov, Zhai, Wightman, Uszkoreit, and Beyer]{DBLP:journals/corr/abs-2106-10270}
Steiner, A., Kolesnikov, A., Zhai, X., Wightman, R., Uszkoreit, J., and Beyer, L.
\newblock How to train your vit? data, augmentation, and regularization in vision transformers.
\newblock \emph{CoRR}, abs/2106.10270, 2021.
\newblock URL \url{https://arxiv.org/abs/2106.10270}.

\bibitem[Tan \& Le(2019)Tan and Le]{tan2019efficientnet}
Tan, M. and Le, Q.
\newblock Efficientnet: Rethinking model scaling for convolutional neural networks.
\newblock In \emph{International conference on machine learning}, pp.\  6105--6114. PMLR, 2019.

\bibitem[Tian et~al.(2020)Tian, Yi, Bai, Tao, Zhang, and Wen]{DBLP:conf/icassp/TianYBTZW20}
Tian, Z., Yi, J., Bai, Y., Tao, J., Zhang, S., and Wen, Z.
\newblock Synchronous transformers for end-to-end speech recognition.
\newblock In \emph{2020 {IEEE} International Conference on Acoustics, Speech and Signal Processing, {ICASSP} 2020, Barcelona, Spain, May 4-8, 2020}, pp.\  7884--7888. {IEEE}, 2020.
\newblock \doi{10.1109/ICASSP40776.2020.9054260}.
\newblock URL \url{https://doi.org/10.1109/ICASSP40776.2020.9054260}.

\bibitem[Touvron et~al.(2021)Touvron, Cord, Douze, Massa, Sablayrolles, and J{\'{e}}gou]{DBLP:conf/icml/TouvronCDMSJ21}
Touvron, H., Cord, M., Douze, M., Massa, F., Sablayrolles, A., and J{\'{e}}gou, H.
\newblock Training data-efficient image transformers and distillation through attention.
\newblock In Meila, M. and Zhang, T. (eds.), \emph{Proceedings of the 38th International Conference on Machine Learning, {ICML} 2021, 18-24 July 2021, Virtual Event}, volume 139 of \emph{Proceedings of Machine Learning Research}, pp.\  10347--10357. {PMLR}, 2021.
\newblock URL \url{http://proceedings.mlr.press/v139/touvron21a.html}.

\bibitem[Vaswani et~al.(2017)Vaswani, Shazeer, Parmar, Uszkoreit, Jones, Gomez, Kaiser, and Polosukhin]{vaswani2017attention}
Vaswani, A., Shazeer, N., Parmar, N., Uszkoreit, J., Jones, L., Gomez, A.~N., Kaiser, {\L}., and Polosukhin, I.
\newblock Attention is all you need.
\newblock \emph{Advances in neural information processing systems}, 30, 2017.

\bibitem[Voita et~al.(2019)Voita, Talbot, Moiseev, Sennrich, and Titov]{DBLP:conf/acl/VoitaTMST19}
Voita, E., Talbot, D., Moiseev, F., Sennrich, R., and Titov, I.
\newblock Analyzing multi-head self-attention: Specialized heads do the heavy lifting, the rest can be pruned.
\newblock In Korhonen, A., Traum, D.~R., and M{\`{a}}rquez, L. (eds.), \emph{Proceedings of the 57th Conference of the Association for Computational Linguistics, {ACL} 2019, Florence, Italy, July 28- August 2, 2019, Volume 1: Long Papers}, pp.\  5797--5808. Association for Computational Linguistics, 2019.
\newblock \doi{10.18653/v1/p19-1580}.
\newblock URL \url{https://doi.org/10.18653/v1/p19-1580}.

\bibitem[Wortsman et~al.(2022)Wortsman, Ilharco, Gadre, Roelofs, Gontijo-Lopes, Morcos, Namkoong, Farhadi, Carmon, Kornblith, and Schmidt]{pmlr-v162-wortsman22a}
Wortsman, M., Ilharco, G., Gadre, S.~Y., Roelofs, R., Gontijo-Lopes, R., Morcos, A.~S., Namkoong, H., Farhadi, A., Carmon, Y., Kornblith, S., and Schmidt, L.
\newblock Model soups: averaging weights of multiple fine-tuned models improves accuracy without increasing inference time.
\newblock In Chaudhuri, K., Jegelka, S., Song, L., Szepesvari, C., Niu, G., and Sabato, S. (eds.), \emph{Proceedings of the 39th International Conference on Machine Learning}, volume 162 of \emph{Proceedings of Machine Learning Research}, pp.\  23965--23998. PMLR, 17--23 Jul 2022.
\newblock URL \url{https://proceedings.mlr.press/v162/wortsman22a.html}.

\bibitem[Xiao et~al.(2022)Xiao, Lin, Seznec, Demouth, and Han]{xiao2022smoothquant}
Xiao, G., Lin, J., Seznec, M., Demouth, J., and Han, S.
\newblock Smoothquant: Accurate and efficient post-training quantization for large language models.
\newblock \emph{arXiv preprint arXiv:2211.10438}, 2022.

\bibitem[Xiao et~al.(2023)Xiao, Tian, Chen, Han, and Lewis]{xiao2023streamingllm}
Xiao, G., Tian, Y., Chen, B., Han, S., and Lewis, M.
\newblock Efficient streaming language models with attention sinks.
\newblock \emph{arXiv}, 2023.

\bibitem[Yin et~al.(2022)Yin, Vahdat, Alvarez, Mallya, Kautz, and Molchanov]{yin2022vit}
Yin, H., Vahdat, A., Alvarez, J.~M., Mallya, A., Kautz, J., and Molchanov, P.
\newblock A-vit: Adaptive tokens for efficient vision transformer.
\newblock In \emph{Proceedings of the IEEE/CVF Conference on Computer Vision and Pattern Recognition}, pp.\  10809--10818, 2022.

\bibitem[Zhang et~al.(2019)Zhang, Kishore, Wu, Weinberger, and Artzi]{zhang2019bertscore}
Zhang, T., Kishore, V., Wu, F., Weinberger, K.~Q., and Artzi, Y.
\newblock Bertscore: Evaluating text generation with bert.
\newblock \emph{arXiv preprint arXiv:1904.09675}, 2019.

\end{thebibliography}
\bibliographystyle{icml2024}

\newpage

\appendix
\section{Related Work} \label{sec:related}
\subsection{Efficient Transformers} \label{subsec:eff_vit}

Notable progress has been made to reduce the high computational cost of neural networks and enable efficient deployment to resource-constrained environments.
At the algorithmic level, methods such as model quantization \cite{DBLP:conf/aaai/ShenDYMYGMK20, DBLP:conf/icml/KimGYMK21,  xiao2022smoothquant}, model pruning \cite{han2015deep, 
DBLP:conf/acl/VoitaTMST19, DBLP:journals/corr/abs-2203-07259}, and knowledge distillation \cite{DBLP:journals/corr/HintonVD15, beyer2022knowledge} have gained popularity. Orthogonal to weight pruning, token compression \cite{yin2022vit, rao2021dynamicvit, liang2022expediting} has shown that transformer inputs can be dynamically pruned at inference time. In this paper, we focus on adaptive token compression techniques, where token reduction decisions are conditioned on the input image.

\subsection{Token Compression}

\textbf{Token Pruning} accelerates the inference of ViT models by discarding less important tokens. Various prior work studies have worked on identifying such token redundancies. DynamicViT \cite{rao2021dynamicvit}, for example, trains a token importance predictor using the Gumbel-Softmax distribution. A-ViT \cite{yin2022vit} learns about the importance of tokens by introducing a loss function that penalizes unpruned tokens. E-ViT \cite{liang2022expediting} uses attention scores from the [CLS] token as the importance heuristic. Although these methods are effective post-deployment, they require costly retraining from scratch or finetuning from a model checkpoint. In contrast, our work focuses on completely bypassing such usability barriers (\cref{tab:comp_app}).


\textbf{Token Merging} combines tokens instead of pruning them. Prior works have attempted to fuse unimportant tokens into a single token using custom heuristics \cite{10.1007/978-3-031-20083-0_37} or learnable MLP projections such as the TokenLearner \cite{DBLP:conf/nips/RyooPADA21}.
Token pooling has also been proposed as a downsampling method via merging \cite{DBLP:journals/corr/abs-2110-03860}; however, its iterative k-means-based method is slow and incompatible with the off-the-shelf models. ToMe \cite{bolya2022tome}, which was recently introduced as a token merging module utilizing a bipartite graph matching algorithm, achieves comparable accuracy to token pruning without any retraining.
Our work makes a case for token merging as a preferred building block for training-free acceleration and makes improvements to push the pareto frontier of the accuracy-efficiency trade-off.

\begin{table}[tb]
    \centering
    \caption{Comparison of different token compression techniques. Our Delayed Spatial Merging (DSM) framework fully embraces training-free acceleration. }
    \label{tab:comp_app}
    \vskip 0.1in
    \resizebox{0.95\columnwidth}{!}{%
    \begin{tabular}{lcccc}
    \toprule
    {}    & Pretrain & Finetune & Training-free \\
    \midrule
    DynamicViT \cite{rao2021dynamicvit}& \xmark  & \cmark & \xmark   \\
    SPViT \cite{kong2021spvit} & \xmark  & \xmark & \xmark   \\
    A-ViT \cite{yin2022vit}      & \xmark  & \cmark & \xmark  \\
    E-ViT \cite{liang2022expediting}      & \cmark  & \cmark & \xmark  \\
    ATS \cite{fayyaz2022ats} & \xmark  & \xmark & \xmark   \\
    ToMe \cite{bolya2022tome}      & \cmark  & \xmark & \cmark$^{\dagger}$  \\
    DSM (\textbf{Ours}) & \xmark & \xmark & \cmark  \\
    \bottomrule
    \end{tabular}
    }
    \begin{flushleft}
\hspace{10pt} \tiny $\dagger$ susceptible to accuracy degradation.
\end{flushleft}
\vskip -0.2in
\end{table}

\subsection{Token Outliers}
To improve the token merging technique, we tackle it from the perspective of a recently observed token outlier problem, which occurs in large transformer models for both vision and language tasks.
Token outliers were popularized in activation quantization research, where certain tokens or channels have much higher activation magnitude than others~\cite{xiao2022smoothquant,dettmers2022llm,lin2023awq,heo2023rethinking}.
Similarly, both supervised and unsupervised ViTs have identified token outliers~\cite{bondarenko2023quantizable,darcet2023registers}, where they are characterized as low-information background tokens that pool global information (similar to the function of the [CLS] token).

The cause of token outliers can be traced back to the Softmax function in attention, where the attention must sum up to one~\cite{softmax1,xiao2023streamingllm}.
When the attention head does not want to update the residual stream, the head executes a ``no-op" by attending heavily to a low-information token~\cite{bondarenko2023quantizable}.
In this work, we confirm that the ``attention sinks" caused by the Softmax function are present--a fact that is subsequently used as a foundation to explore the unique attention behavior in ViTs. 

\subsection{Training-Free Acceleration}
For ViTs, most of the models used in classification tasks are small (or tiny) variants in the ViT and DeiT model families.
Prior training-based token reduction techniques have experimented with a focus on small models due to the high training cost of larger models~\cite{rao2021dynamicvit,liang2022expediting}.
When considering the training and hyperparameter tuning costs, the total computations can become unwieldy for many researchers and practitioners \cite{DBLP:journals/corr/abs-2106-10270}.
Motivated by the fact that the highest-performing models are too expensive to compress, we propose \textbf{a training-free framework for compressing large ViTs}.
We take the method's speed as an equally important figure of merit as the final model performance and constrain our solution to be training-free. Our work addresses the following question: How do we compress ViTs without expensive training to realize high-accuracy inference models?

\section{Detailed Methodology}
\label{appdx:method}
\begin{figure*}[!h]
\centering
\includegraphics[width=0.7\paperwidth]{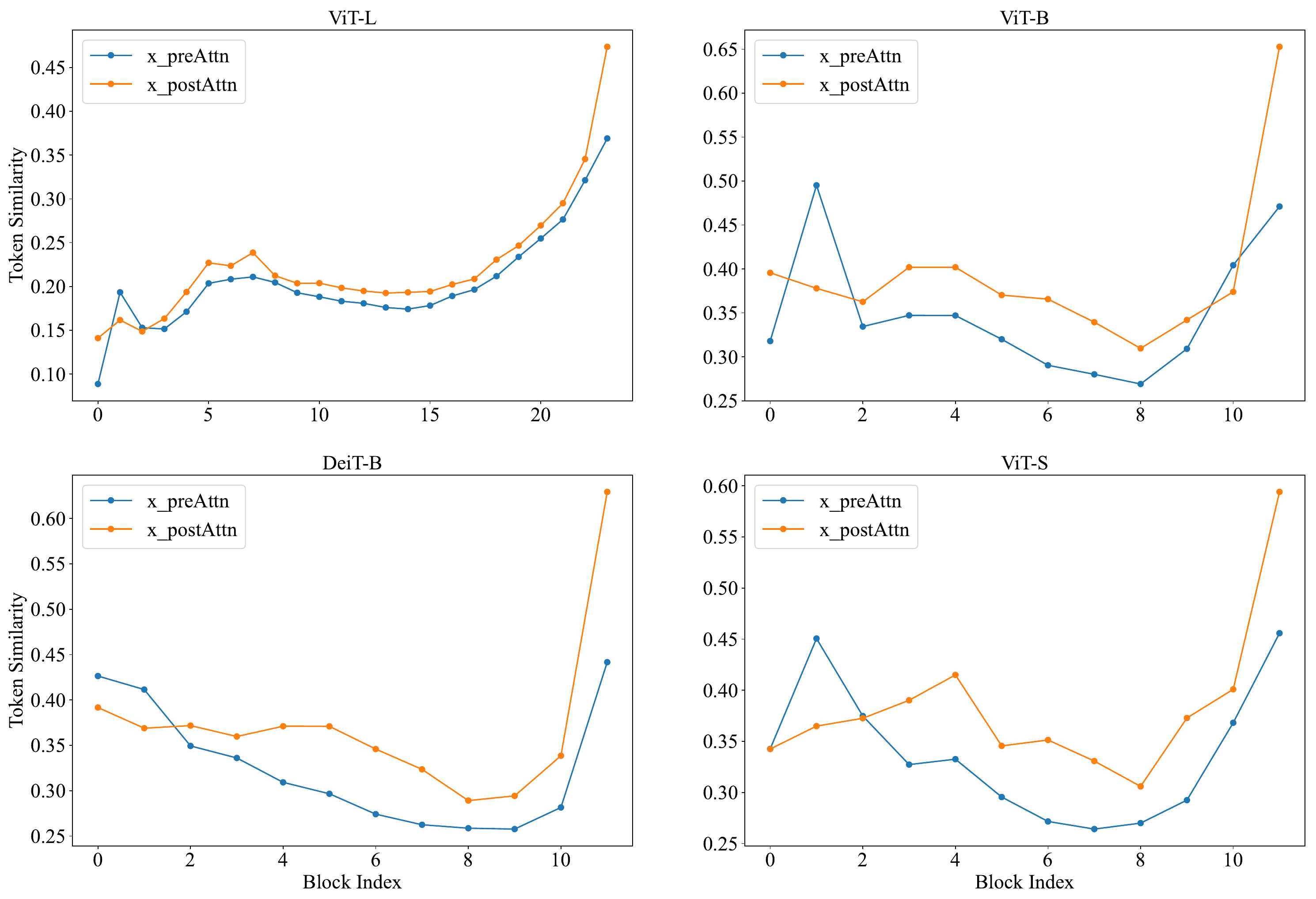}
\caption{\textbf{Delayed convergent attention} phenomena is observed for various pretrained visual transformers. Attention block consistently makes the tokens more similar after a certain threshold layer, which is around 1/6th of the network.}
\label{fig:appdx:cossim}
\end{figure*}

\begin{figure*}[!h]
\centering
\includegraphics[width=0.6\paperwidth]{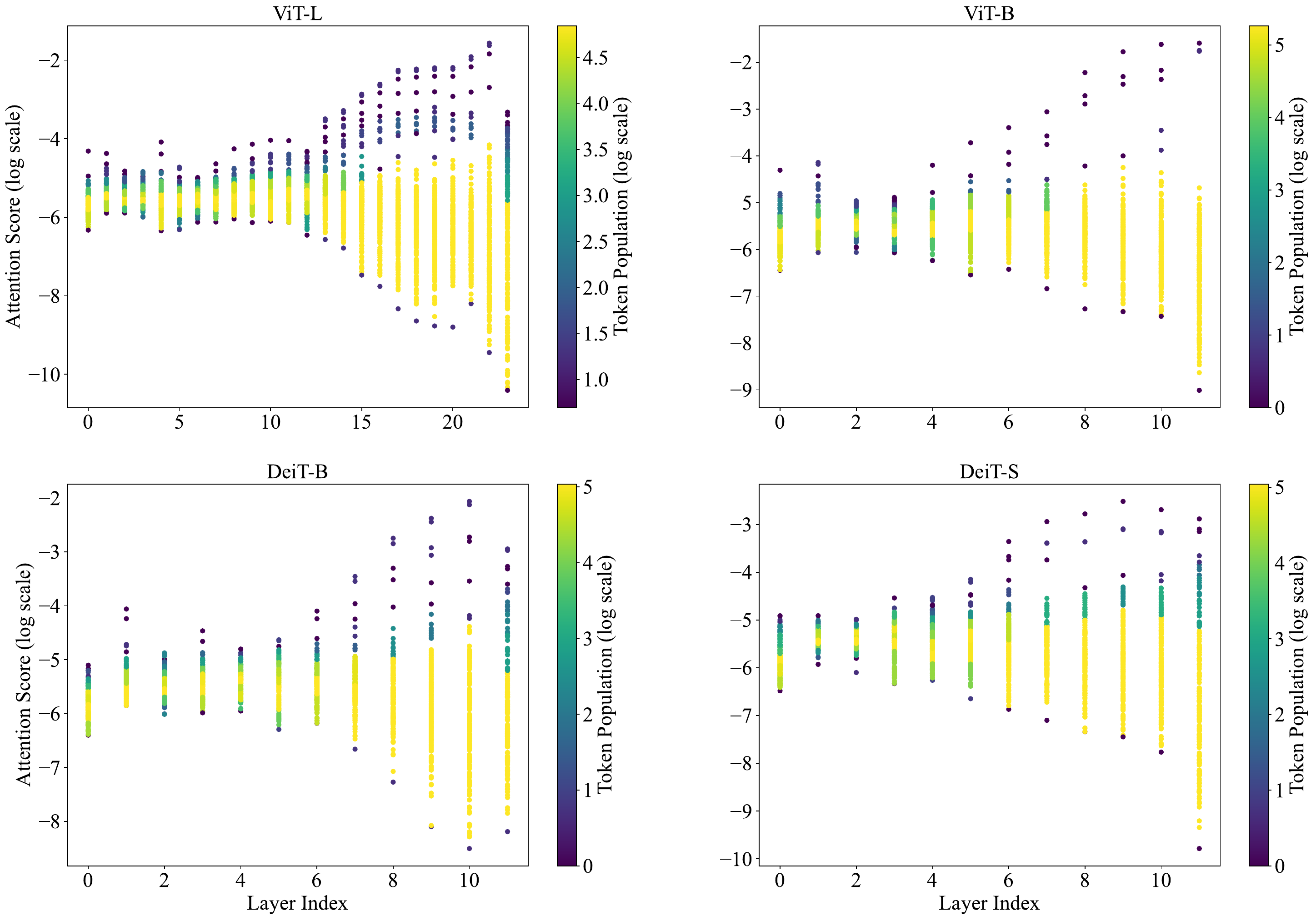}
\caption{Outlier tokens observed in different ViT architectures.}
\label{fig:appdx:attn-quant}
\end{figure*}

\paragraph{DSM Hyperparameters.}
We fix the delay parameter $D$ to be the transition point where the convergent attention behavior begins to emerge. That is, for a DeiT-S model with a depth of 12, we choose $D=2$ as the convergent attention appears in the second block (ref. \cref{fig:show}. We visualize additional networks in~\cref{fig:appdx:cossim} and \cref{fig:appdx:attn-quant}, where the 1/6th point of the network is generally the point at which the attention behavior switches from divergent to convergent.

The localized merging parameter $T$ can be fixed as a function of the window size $w$ and the reduction rate $r$. This is because localized merging with progressively increasing window size naturally degenerates into global merging. With gradual token merging, the sequence length becomes smaller than the window size itself. Thus, increasing the window size yields a partial localized merging that smoothly transitions to global merging.

\section{More Experiments}
\subsection{The Case for Merging}
Before delving into our DSM evaluation, we first make a more fundamental case that token merging is the right building block over token pruning for training-free ViT acceleration.
Off-the-shelf ViT models are commonly pretrained with a dense token distribution (no token dropping).
Thus, a trained model ``expects'' to see not only task-relevant tokens but also less relevant ones like background tokens.
Less informative tokens can also function as regularization, which makes it risky to assume that less important tokens can be removed without degrading the prediction performance.
Thus, token pruning may not be the optimal design choice for the training-free setting.


\textbf{Heuristics Ablation for Vanilla Token Compression.}
As in~\cref{tab:token_merg_vs_drop}, we empirically support the case for token merging by comparing pruning to merging with various importance criteria. For pruning, the lowest L2 norm is dropped; for merging, the highest cosine similarity score is merged. We observe that the output of attention block $X$ is a surprisingly good heuristic for pruning, but it lags merging options by 2\%. The $K$ embedding criteria yield the highest performance for merging. It best represents the tokens, even more than $X$, which has a larger embedding size per token. This may be due to of overparameterized embeddings, where having more channels can result in noise. Since $K$ has less number of channels through the multi-head attention, its compact representation can resolve this problem. 

\textbf{Heuristics Ablation for DSM.}
We ablate the best similarity heuristic for our framework. As shown in \cref{tab:our-heuristic}, $K$ is the best choice. Random selection has the worst accuracy while choosing any other heuristic leads to a worst accuracy-throughput trade-off. 

\begin{table}[th!]
\caption{\textbf{Prune vs. Merge} comparison using ViT-L with $r=7$. Merging retains accuracy more effectively in training-free settings. X is inside the attention block.}
\label{tab:token_merg_vs_drop}
\begin{center}
\begin{sc}
\begin{tabular}{lcc|cc}
\toprule
{} &  \multicolumn{2}{c}{Prune} & \multicolumn{2}{c}{Merge} \\
\midrule
Features    & acc & im/s & acc & im/s \\
\midrule
Random    & 2.96             & 131.3  & 61.89  & 136.5  \\
X & \textbf{81.58}          & 129.1  & 83.41 & 125.7  \\
K         & 71.86           & \textbf{130.7}  & \textbf{83.51}  & \textbf{132.7}  \\
Q         & 73.65           & 130.3 & 83.25 & 131.3  \\
V         & 78.9            & 130.3 & 83.44 & 132.3  \\
\bottomrule
\end{tabular}
\end{sc}
\end{center}
\end{table}

\begin{table}[th!]
    \centering
    \caption{\textbf{Design Choices.} When DSM is applied to ViT-L with $r=18$, the K embeddings yield the best accuracy-throughput trade-off.}
    \label{tab:our-heuristic}
    \vskip 0.1in
    \begin{sc}
    \resizebox{\columnwidth}{!}{%
    \begin{tabular}{lcccccc}
    \toprule
    {}       & Random & X & K & Q & V \\
    \midrule
    acc & 59.6 & 82.7 & \textbf{82.9} & 82.4 & 82.6 \\
    im/s & 589.1 & 552.2 & \textbf{591.5} & 589.2 & 589.5 \\
    \bottomrule
    \end{tabular}
    }
\end{sc}
\end{table}

\begin{figure}[t!]
     \centering
     \includegraphics[width=.9\columnwidth]{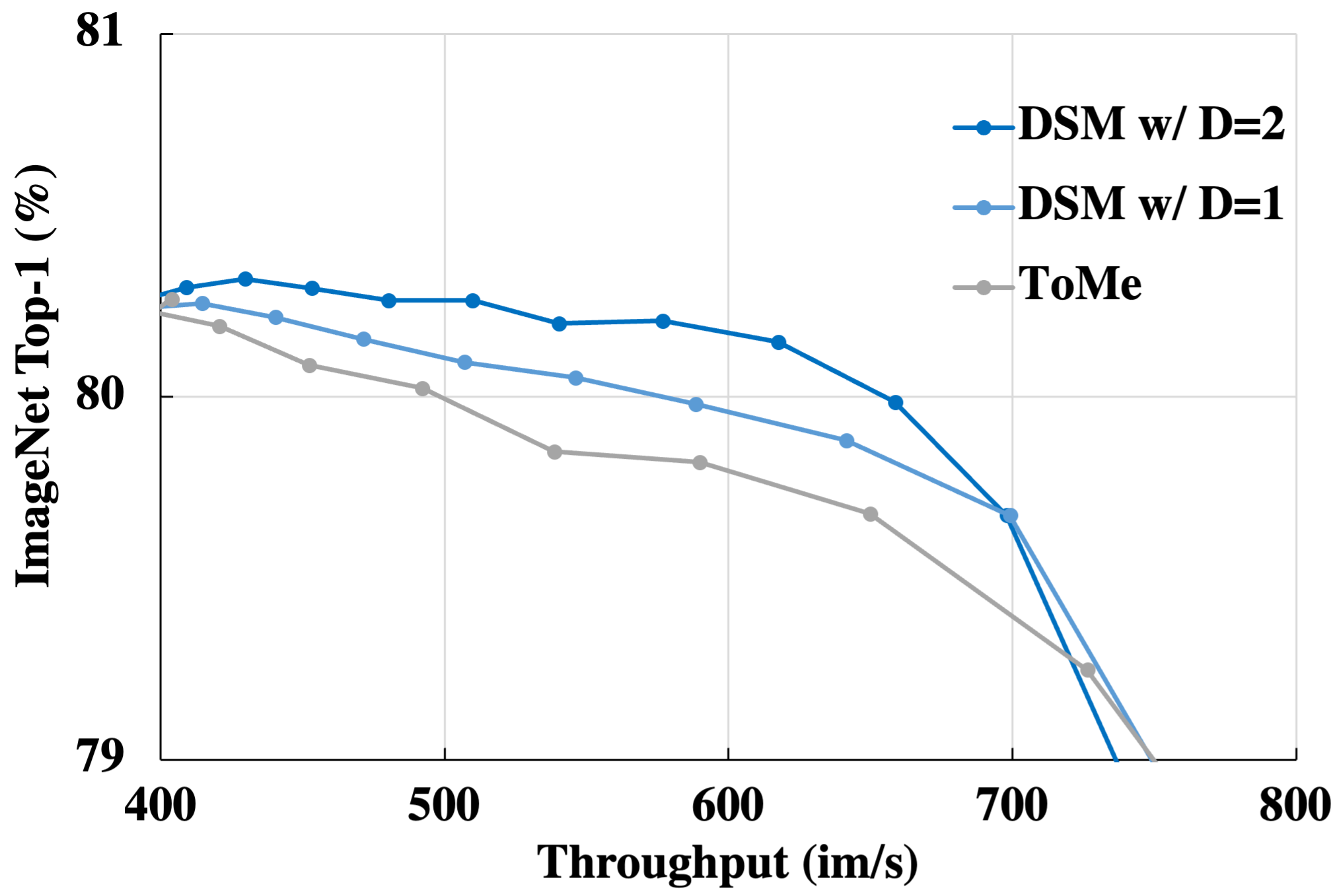}
     \caption{\textbf{Sharpness-minimized Models} trained with the SAM optimizer on ViT-B are more friendly to compression. It allows 1.6$\times$ throughput gain with the help of delayed merging (denoted as $D$).}
     \vspace{-3mm}
     \label{fig:sam-thp}
\end{figure}

\begin{figure}[!]
     \centering     \includegraphics[width=0.85\columnwidth]{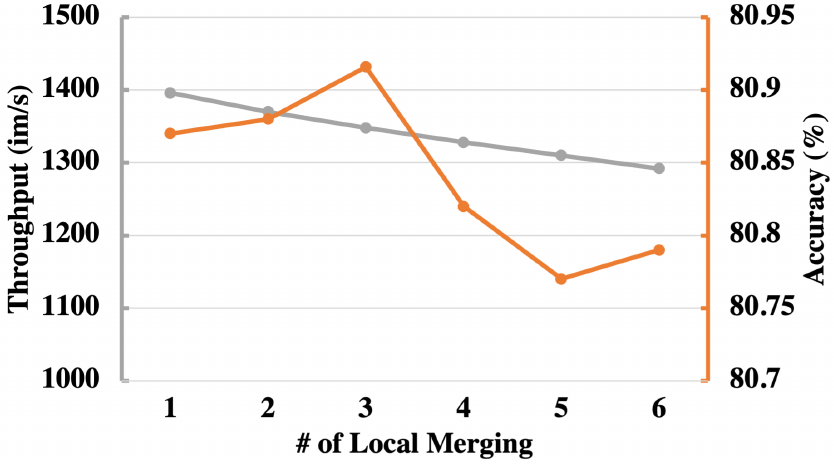}
     \caption{Effect of localized merging on throughput and accuracy. We observe that higher use of localized merging leads to less throughput due to the additional data movements. Yet, there is a general sweet spot for $L=3$, which is the default setting we found using the increasing window technique.}
     \label{fig:local_ablation}
\end{figure}

\subsection{Main results}
We also evaluate the efficacy of DSM against training-based acceleration in various vision architectures that are not transformer-based. CNNs are known to be more parameter-efficient due to the weight-sharing nature of convolutions and smaller peak memory (since it does not use quadratic self-attention). As in~\cref{tab:eff-arch}, we observe that DeiT-S w/ DSM performs comparably in the accuracy-compute trade-off with much more expensive methodologies such as EfficientNet via Neural Architecture Search~\cite{tan2019efficientnet}. 

\begin{table}[!t]
\centering
\caption{Comparison to convolution-based vision architectures.}
\label{tab:eff-arch}
\resizebox{\columnwidth}{!}{
\begin{tabular}{lcccc}
    \toprule
    Model       & Top-1 & Speedup ($\uparrow$) \\
    \midrule
    DeiT-S  & 79.8  & 1$\times$   \\
    \midrule
    EfficientNet-B2~\cite{tan2019efficientnet}  & 80.1  & 1.33$\times$   \\
    EfficientNet-B3~\cite{tan2019efficientnet}  & 81.6  & 0.78$\times$   \\
    ResNet-152~\cite{he2016deep}  & 78.3 & 0.56$\times$   \\
    RegNetY-4GF~\cite{radosavovic2020regnet}  & 80.0  & 1.23$\times$   \\
    \midrule
    DeiT-S w/ DSM ($r$16)  & 79.6  & 1.5$\times$   \\
    DeiT-S w/ DSM ($r$18)  & 79.4  & 1.6$\times$   \\
    \bottomrule
\end{tabular}
}
\end{table}

Moreover, we conduct additional experiments for both larger (ViT-H) and smaller (DeiT-Ti) models. Table \ref{tab:combined_compare} compares DSM against ToMe using ViT-H and DeiT-T, respectively. 

\begin{table}[th!]
    \centering
    \caption{Comparison of ViT-H and DeiT-T @ ImageNet-1k}
    \label{tab:combined_compare}
    \vskip 0.1in
    \begin{sc}
    \resizebox{\columnwidth}{!}{%
    \begin{tabular}{lcccc}
    \toprule
    \multirow{7}{*}{\rotatebox{90}{\textbf{ViT-H }}} & {} &  $\Delta$ Accuracy(\%) & Throughput (image/sec) & \# MACs (G) \\
    \cmidrule{2-5}
    & ToMe & -0.2  & 50.18 & 145.84 \\
    & DSM & -0.2 & \textbf{56.90} &  \textbf{129.64} \\
    \cmidrule{2-5}
    & ToMe & -0.6 & 64.01 &  113.90 \\
    & DSM & -0.6 & \textbf{72.60} &  \textbf{101.79} \\
    \cmidrule{2-5}
    & ToMe & -0.8 & 70.38 &  103.36 \\
    & DSM & -0.8 & \textbf{79.30} &  \textbf{92.59} \\
    \midrule
    \multirow{6}{*}{\rotatebox{90}{\textbf{DeiT-T }}} & ToMe & -0.1  & 2457 & 1.18 \\
    & DSM & -0.1 & \textbf{2722} &  \textbf{1.09} \\
    \cmidrule{2-5}
    & ToMe & -0.5 & 3020 &  0.93 \\
    & DSM & -0.5 & \textbf{3257} &  \textbf{0.86} \\
    \cmidrule{2-5}
    & ToMe & -2.0 & 3881 &  0.69 \\
    & DSM & -2.0 & \textbf{4001} &  \textbf{0.71} \\
    \bottomrule
    \end{tabular}
    }
    \end{sc}
\end{table}

\begin{figure*}[!t]
\centering
\includegraphics[width=0.75\paperwidth]{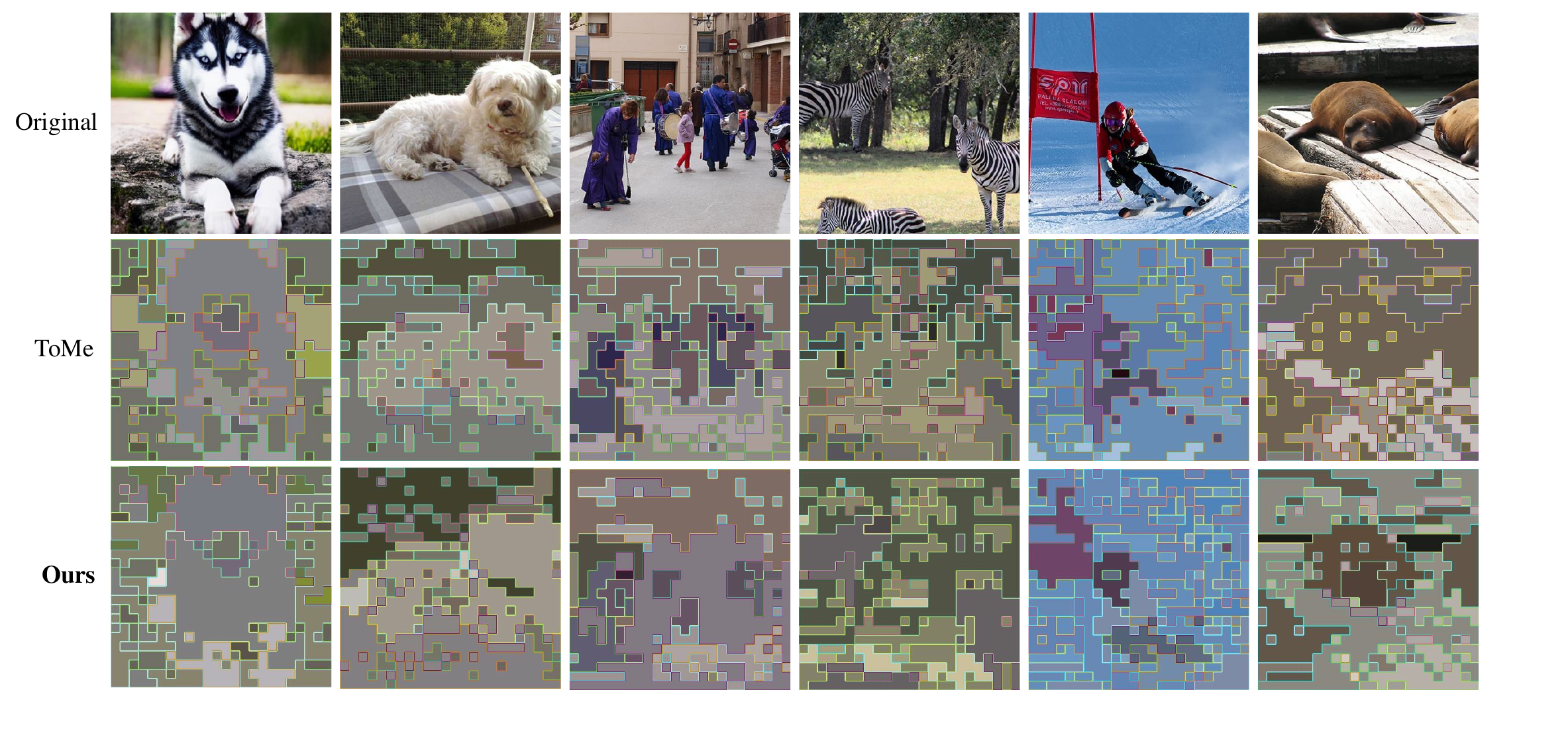}
\caption{Qualitative comparison of DSM to ToMe using a ViT-L$_{384}$ model.
Merged tokens share the same border and filling color. 
DSM merges more contiguous patches that are semantically similar, leading to more interpretable outcomes that retain the original features.}
\label{fig:visuals}
\end{figure*}
\section{More Visualizations}
In \cref{fig:visuals}, we show the input tokens belonging to the final merged token. We use $r=24$ for ToMe and $r=28$ with $D=4$ and $w=8$ for our framework. Note that the parameters are different since the resolution is higher. To match the final token count, we do not merge the last block in our framework. We see that in the second image, the face of the Maltese is contiguously merged into a single token for us, while ToMe separates out the nose. The same is true for the body of a Huskey in the first photo and the people in the center of the third photo, where our framework tends to merge more contiguous tokens. 


\begin{figure*}[th]
\centering
\includegraphics[width=0.7\paperwidth]{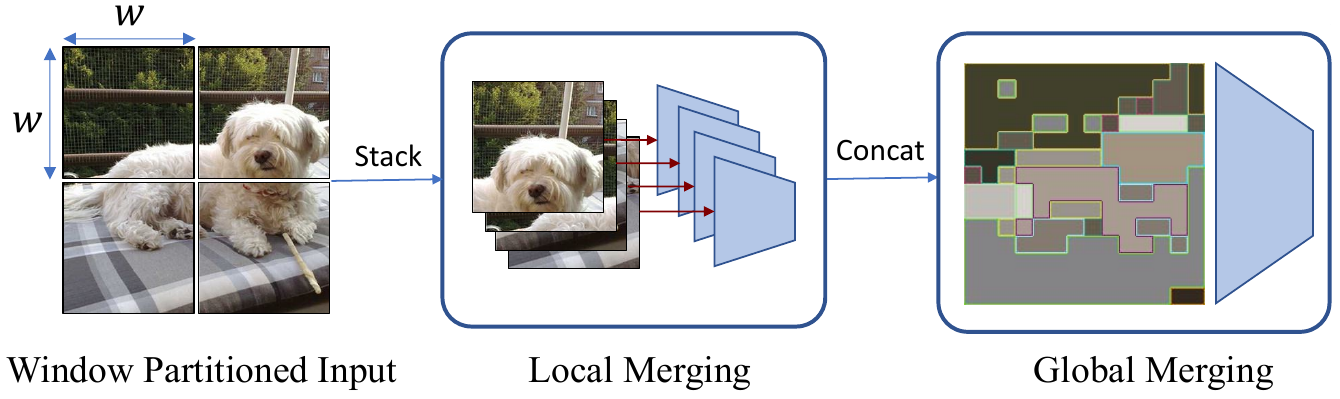}
\caption{Local merging with window partitioning is illustrated with a visual input.}
\label{fig:appdx:local}
\end{figure*}

\end{document}